\documentclass[letterpaper, 10 pt, conference]{ieeeconf}  

\IEEEoverridecommandlockouts                              

\usepackage{graphics} 
\usepackage{epsfig} 
\usepackage{mathptmx} 
\usepackage{times} 
\usepackage{amsmath} 
\usepackage{amssymb}  
\usepackage{todonotes}
\usepackage{etoolbox}

\usepackage[font=small,labelfont=bf]{caption} 

\usepackage{hyperref}

\usepackage{sidecap}

\title{\LARGE \bf
kPAM-SC: Generalizable Manipulation Planning using \\ KeyPoint Affordance and Shape Completion
}

\author{Wei Gao, Russ Tedrake 
\thanks{CSAIL, Massachusetts Institute of Technology, 77 Massachusetts Ave, Cambridge, USA. Emails: weigao@mit.edu, russt@mit.edu.}
}

\begin{document}

\maketitle
\thispagestyle{empty}
\pagestyle{plain}

\begin{abstract}

Manipulation planning is the task of computing robot trajectories that move a set of objects to their target configuration while satisfying physically feasibility. In contrast to existing works that assume known object templates, we are interested in manipulation planning for a category of objects with potentially unknown instances and large intra-category shape variation. To achieve it, we need an object representation with which the manipulation planner can reason about both the physical feasibility and desired object configuration, while being generalizable to novel instances. The widely-used pose representation is not suitable, as representing an object with a parameterized transformation from a fixed template cannot capture large intra-category shape variation. 
Hence building on our previous work kPAM\footnote{kPAM~\cite{manuelligao2019kpam} stands for \textbf{K}ey\textbf{P}oint \textbf{A}ffordance-based \textbf{M}anipulation.}, we propose a new hybrid object representation consisting of semantic keypoint and dense geometry (a point cloud or mesh) as the interface between the perception module and motion planner. Leveraging advances in learning-based keypoint detection and shape completion, both dense geometry and keypoints can be perceived from raw sensor input. Using the proposed hybrid object representation, we formulate the manipulation task as a motion planning problem which encodes both the object target configuration and physical feasibility for a category of objects. In this way, many existing manipulation planners can be generalized to categories of objects, and the resulting perception-to-action manipulation pipeline is robust to large intra-category shape variation. Extensive hardware experiments demonstrate our pipeline can produce robot trajectories that accomplish tasks with never-before-seen objects. The video demo and source code are available on our project page: \href{https://sites.google.com/view/generalizable-manipulation/home}{\textcolor{blue}{\underline{https://sites.google.com/view/generalizable-manipulation}}}.

\end{abstract}

\section{Introduction}
\label{sec:intro}

This paper focuses on robotic manipulation planning at a category level. In particular, the pipeline should take raw sensor inputs and plan physically feasible robot trajectories that will move a set of objects to their target configuration. For example, the robot should ``place mugs in a box" and handle potentially unknown instances in the mug category, despite the variation of shape, size, and topology. Accomplishing these type of tasks is of significant importance to both industrial applications and interactive assistant robots.

To achieve this goal, we need an object representation that (1) is accessible from raw sensor inputs, (2) with which the manipulation planner can reason about both the physical feasibility and desired object configuration, (3) can generalize to novel instances. Perhaps 6-DOF pose is the most widely-used object representation in robotic manipulation. Most manipulation planning algorithms~\cite{toussaint2018differentiable, schulman2013finding, garrett2018sampling} assume the known geometric template and 6-DOF pose of the manipulated objects.
Consequently, many contributions from the vision community~\cite{sahin2018category, wang2019normalized, myronenko2010cpd, gao2018filterreg} focus on pose estimation from raw sensor observations. 
However, as detailed in Sec.~\ref{subsec:sc_result}, representing an object with a parameterized pose defined on a fixed geometric template, as these works do, may not adequately capture large intra-class shape or topology variations. Directly using these poses for manipulation can lead to physical infeasibility for certain instances in the category. On the other hand, a large body of work~\cite{gualtieri2016gpd, morrison2018closing, zeng2018affordance, kalashnikov2018qt} focuses on grasping planning. Although these methods enable robotic picking for arbitrary objects, it is hard to extend them to more complex tasks that involve object target configuration.


In our previous work kPAM~\cite{manuelligao2019kpam}, we represent objects using semantic 3D keypoints, which provides a concise way to specify the target configuration for a category of objects. Although this approach has successfully accomplished several category-level manipulation tasks, it lacks the complete and dense geometric information of the object. Thus, kPAM~\cite{manuelligao2019kpam} cannot reason about physical properties such as the collision, static equilibrium, visibility, and grasp stability of the planned robot action, despite their practical importance in robotic applications.
As a result, authors of kPAM~\cite{manuelligao2019kpam} need to manually specify various intermediate robot configurations to ensure physical feasibility, which can be labor-intensive and sub-optimal.

Building on kPAM~\cite{manuelligao2019kpam}, we resolve this limitation with a new hybrid object representation which combines both (i) semantic 3D keypoints and (ii) full dense geometry (a point cloud or mesh).
The dense geometry is obtained by leveraging well-established shape completion algorithms~\cite{zhang2018learning, mescheder2019occupancy, park2019deepsdf}, which generalize well to novel object instances. With the combined dense geometry and keypoints as the object representation, we formulate the manipulation task as a motion planning problem that can encode both the object target configuration and physical feasibility for a category of objects. This motion planning problem can be solved by a variety of existing planners and the resulting robot trajectories can move the objects to their target configuration in a physically feasible way. The entire perception-to-action manipulation pipeline is robust to large intra-category shape variation. Extensive hardware experiments demonstrate our method can reliably accomplish manipulation tasks with never-before-seen objects in a category.

The contribution of this work is twofold. Conceptually, we introduce a hybrid object representation consists of dense geometry and keypoints as the interface between the perception module and planner. This representation has similar functionalities with existing 6-DOF pose representation with templates, while the generalizability to novel instances makes it a promising alternative. On the perspective of implementation, we contribute a novel integration of shape completion with the keypoint detection and manipulation planning. This integration enables many existing manipulation planners, either optimization-based methods~\cite{schulman2013finding, toussaint2018differentiable} or sampling-based approaches~\cite{garrett2018sampling, srivastava2014combined}, to handle a category of objects in a unified and precise way.

This paper is organized as follows: in Sec.~\ref{sec:related_works} we review related works. Sec.~\ref{sec:kpam} provides background knowledge about the keypoint representation in~\cite{manuelligao2019kpam}. Sec.~\ref{sec:pipeline} describes our manipulation pipeline. Sec.~\ref{sec:results} demonstrates our methods on several manipulation tasks and shows generalization to novel instances. Sec.~\ref{sec:conclusion} concludes this work.

\section{Related Work}
\label{sec:related_works}

\subsection{Pose-based Robotic Manipulation}

For most manipulation applications, the object pose is the default interface between the perception and planning modules: the perception module estimates the pose from raw sensor inputs; the planning module takes the estimated pose as input and plans robot actions to accomplish some manipulation tasks. Researchers have made many contributions on both the pose estimation~\cite{sahin2018category, wang2019normalized, myronenko2010cpd, gao2018filterreg} and manipulation planning~\cite{schulman2013finding, toussaint2018differentiable, garrett2018sampling, srivastava2014combined}. Some works~\cite{chitta2012perception, kappler2018real} integrate state-of-the-art pose estimators with trajectory planners to build fully functional manipulation pipelines and solve real-world tasks such as packing and assembly.

However, pose estimation can be ambiguous under large intra-category shape variations, and using one geometric template for motion planning and object target specification can lead to physically infeasible states for other instances within a category of objects. Thus, we need a more generalizable object representation as the interface between the perception and planning modules, the one consists of dense geometry and semantic keypoint proposed in this work would be a promising candidate.

\subsection{Manipulation at a Category Level}

Several existing works aim at robot manipulation for a category of object. Among them, kPAM~\cite{manuelligao2019kpam} uses a modularized pipeline where the semantic keypoints are used to represent the object.
However as mentioned in Sec.~\ref{sec:intro}, kPAM cannot reason about physical feasibility. This limitation is resolved in this work by the integration of shape completion and manipulation planning.
%
On the other hand, ~\cite{gualtieri2018pick} demonstrates robotic pick-and-place manipulation across different instances using end-to-end reinforcement learning. However, the reward engineering and training procedure required in these algorithms make it hard to generalize to new target configurations and tasks.

\subsection{Grasping and Manipulation with Shape Completion}

Robot grasp planning is the task of computing a stable grasp pose that allows the robot to reliably pick up the object. Among various approaches for grasp planning, model-free methods~\cite{zeng2018affordance, gualtieri2016gpd, morrison2018closing, mahler2019learning} typically train deep networks that take raw sensor observations as input and produce grasp poses as output. In contrast, model-based algorithms~\cite{zeng2017multi, mahler2016dex, varley2017shape} estimate the grasp quality based on geometric information such as antipodal points or surface normal. 

As the geometry obtained from typical RGBD sensors are noisy and incomplete, several works~\cite{varley2017shape, lundell2019robust, watkins2018multi} combine shape completion with grasping planning for improved performance and robustness. \cite{price2019inferring} also shows shape completion can improve the performance of robot object searching.

In this work, we are interested in category-level robotic manipulation planning. This task requires reasoning about both the desired object configuration and physical feasibility, and is out of the scope for the above-mentioned methods.

\begin{figure}[t]
\centering
\includegraphics[width=0.48\textwidth]{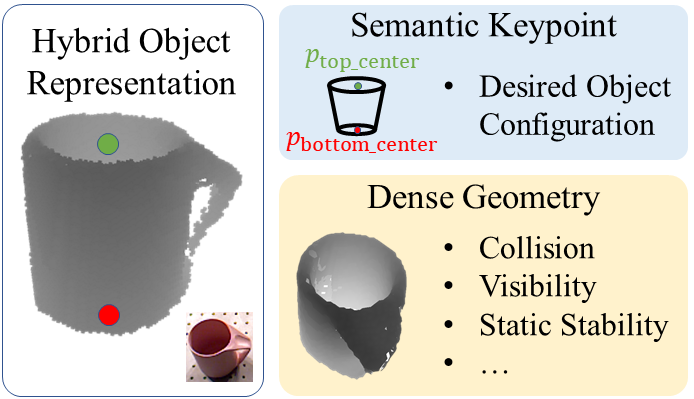}
\caption{\label{fig:kpam} An overview of the proposed hybrid object representation for category-level manipulation planning using the mug as an example. We exploit a hybrid object representation consists of semantic keypoints and dense geometry. The semantic keypoint are used to specify the desired object target configuration, while the dense geometry is used to ensure the physical feasibility of the planned robot action. Benefit from advances in learning based keypoint detection and shape completion, the proposed object representation can be obtained from raw images, and the resulting perception-to-action pipeline generalizes to novel instances within the given category. }
\end{figure}

\begin{figure*}[t]
\centering
\includegraphics[width=0.9\textwidth]{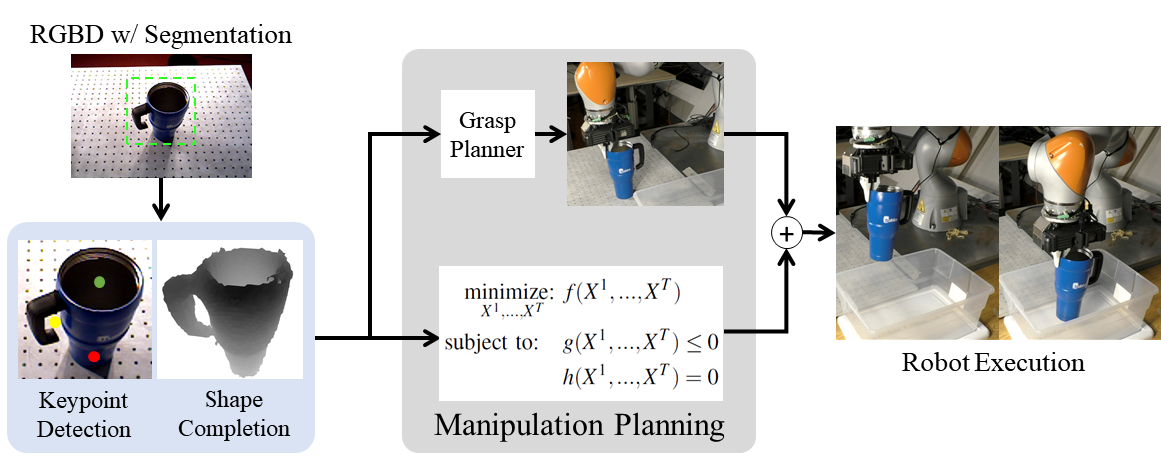}
\caption{\label{fig:pipeline} An overview of the manipulation pipeline. The hybrid object representation consists of dense geometry and keypoint is used as the interface between the perception module and manipulation planner. Given a RGBD image with instance segmentation, we perform shape completion and keypoint detection to obtain dense geometry and 3D keypoints, respectively. Then, the perception result is used to plan robot trajectories that move the objects to their desired configurations while satisfying physical constraints. Semantic keypoints are used to specify the object target configuration, while dense geometry is used to ensure the physical feasibility of the planned robot action. }
\end{figure*}

\section{Preliminary: kPAM}
\label{sec:kpam}

In this section, we give a brief recap kPAM~\cite{manuelligao2019kpam}. kPAM is a framework to specify the object target configuration for robotic manipulation. In kPAM, each object is represented by a set of semantic 3D keypoints $p \in R^{3 \times N}$, where $N$ is the number of keypoints. Using the mug category as an example, we may represent the mug by $N=2$ keypoints: $p_\text{top\_center}$ and $p_\text{bottom\_center}$, as shown in Fig.~\ref{fig:kpam}. These keypoints are detected from raw sensor inputs using state-of-the-art learning-based detectors.

In kPAM, the robot action is abstracted as a rigid transformation $T_\text{action} \in SE(3)$ on the object, and the transformed keypoint would be $T_\text{action} p$.  The manipulation target is defined as a set of geometric costs/constraints on the transformed keypoint $T_\text{action} p$. Using the mug in Fig.~\ref{fig:kpam} as an example, to place the mug at some target location $p_\text{target}$, the planned robot action $T_\text{action}$ should satisfy
\begin{equation}
\label{equ:mug_target_position}
||T_\text{action} p_\text{bottom\_center} - p_\text{target}|| = 0
\end{equation}
After the object has been grasped, the $T_\text{action}$ would be applied to the object by applying the $T_\text{action}$ to the robot end-effector, as the object is assumed to be rigidly attached to the gripper after grasping.


As the dense geometry information of the object is missing, it remains unclear how to plan robot trajectories that apply the $T_\text{action}$ in a physically feasible way. In this work, we achieve this by the integration of dense geometry from shape completion.

\section{Manipulation Pipeline}
\label{sec:pipeline}

As illustrated in Fig.~\ref{fig:pipeline}, we use the hybrid object representation consists of dense geometry and keypoints as the interface between the perception and planning modules. The semantic keypoints are designated manually and used to specify the object target configuration, while the dense geometry is used to ensure the physical feasibility of the planned robot action. The perception part includes shape completion and keypoint detection, which is detailed in Sec.~\ref{subsec:perception}. The manipulation planning and grasp planning that use the perceived results are presented in Sec.~\ref{subsec:planning} and Sec.~\ref{subsec:grasping}, respectively.

\subsection{Perception}
\label{subsec:perception}

The task of perception is to produce the proposed hybrid object representation from raw sensor inputs, which consists of 3D keypoints and dense geometry (point cloud or mesh).
For 3D keypoints we adopt the method in our previous work kPAM~\cite{manuelligao2019kpam}, and this subsection would mainly focus on the perception of dense geometry. For dense geometry, we leverage recent advances in shape completion to obtain a complete point cloud or mesh of the object. Note that although we present specific approaches used in our pipeline, any technique for keypoint detection and shape completion can be used instead.

We use the state-of-the-art ShapeHD network~\cite{zhang2018learning} for 3D shape completion. ShapeHD is a fully convolutional network that takes RGBD images as input and predicts 3D volumetric occupancy. Then the completed point cloud can be extracted by taking the occupied voxel. If the object mesh is required, triangulation algorithms such as marching cubes can be used. The completed geometry are aligned with the observed object (viewer-centered in~\cite{zhang2018learning}) and expressed in the camera frame, we can further transform it into the world frame using the calibrated camera extrinsic parameters.

The shape completion network requires training data consists of RGBD images and corresponding ground-truth 3D occupancy volumes. We collect training data using a self-supervised method similar to LabelFusion~\cite{marion2017labelfusion}. Given a scene containing one object of interest we first perform 3D reconstruction of that scene. Then, we perform background subtraction to obtain the reconstructed mesh of the object. Finally, we can get the occupancy volume by transforming the reconstructed mesh into camera frame and voxelization. Note that the data generation procedure does not require pre-built object template or human annotation. In our experiment, we scan 117 training scenes and collect over 100,000 pairs of RGBD images and ground-truth 3D occupancy volumes within four hours. Even with small amount of data we were able to achieve reliable and generalizable shape completion, some qualitative results are shown in Sec.~\ref{subsec:sc_result}.

\subsection{Manipulation Planning}
\label{subsec:planning}

The manipulation planning module produces the robot trajectories given the perception result. As mentioned in Sec.~\ref{sec:pipeline}, we represent an object $o_j$ by its 3D semantic keypoints $p_j \in R^{3 \times N}$ and dense geometry (point cloud or mesh), where $1 \leq j \leq M$ and $M$ is the number of objects. 

Following many existing works~\cite{hauser2010task, garrett2018sampling, srivastava2014combined}, we assume that the robot can change the state of an object only if the object is grasped by the robot. Furthermore, we assume the object is rigid and the grasp is tight. In other words, there is no relative motion between the gripper and the grasped object. Both the semantic keypoints and dense geometry would move with the robot end-effector during grasping. To achieve this, we need a grasp planner which is discussed in Sec.~\ref{subsec:grasping}.

Given the object representation, the concatenated configuration for robot and objects at time $t$ is defined as $X^t=[o^t_1,...,o^t_M, q^t]$, where $1 \leq t \leq T$, $T$ is the number of time knots and $q^t$ is the robot joint configuration. The general planning problem can be written as
\begin{align}
\label{equ:problem}
    \underset{X^1, ..., X^T}{\text{minimize: }}& f(X^1,...,X^T) \\
    \text{subject to:}~~~&g(X^1,...,X^T) \leq 0 \\
    & h(X^1,...,X^T) = 0
\end{align}
\noindent where $f$ is the objective function, $g$ and $h$ are the concentrated inequality and equality constraints. If optimization-based planning algorithms~\cite{schulman2013finding, drake} are used to solve the problem (\ref{equ:problem}), $f$, $g$ and $h$ should be differentiable. On the other hand, many sampling-based planners~\cite{lavalle1998rapidly, karaman2011sampling} or TAMP algorithms~\cite{garrett2018sampling, kaelbling2010hierarchical} only need a binary predicate on whether the constraint is satisfied.

Using the proposed object representation consist of semantic 3D keypoints and dense geometry, the key benefit is that the motion planner can handle a category of objects without instance-wise tuning. In the following text, we discuss several important costs and constraints that are related to the object representation.

\vspace{0.3 em}
\noindent \textbf{Object Target Configuration} Let $p^t_j$ be the keypoints of the object $o_j$ at the time $t$, where $1 \leq t \leq T$. The target configuration of an object $o_j$ can be represented as a set of costs and constraints on its semantic keypoint $p^T_j$, where $T$ is the terminal time knot. For instance, to place the mug at some target location $p_\text{target}$ as illustrated in Fig.~\ref{fig:kpam}, we need an equality constraint
\begin{equation}
    ||p^T_\text{bottom\_center} - p_\text{target}|| = 0
\end{equation}
\noindent where $p^T_\text{bottom\_center}$ is the mug bottom-center keypoint expressed at time $T$. Note that this constraint can handle mugs with different size, shape and topology. Many other costs and constraints can be used to specify the object target configuration. Please refer to kPAM~\cite{manuelligao2019kpam} for more details.

\vspace{0.3 em}
\noindent \textbf{Collision Avoidance} The dense geometry information from shape completion can be used to ensure the planned trajectory is collision-free. Specifically, let $B_{r}$ denote the set of rigid bodies of the objects, robot and environment, where the geometry of objects are obtained using shape completion. We need to ensure
\begin{align}
    \text{signed\_distance}(X^t; b_{i}, b_{j}) \geq \delta_\text{safe} \\
    \text{ for } b_i \in B_r,~b_j \in B_r,~i \neq j,~1 \leq t \leq T
\end{align}

\noindent where $\delta_\text{safe}$ is a threshold, $\text{signed\_distance}(X^t; b_{i}, b_{j})$ is the signed distance~\cite{schulman2013finding} between the pair of rigid body $(b_i, b_j)$ at the configuration $X^t$. Practically, it is usually unnecessary to check the collision of every rigid body pairs, as most rigid bodies except the grasped object and robot end-effector have limited movement.

\vspace{0.3 em}
\noindent \textbf{Geometric Predicates} In many planning algorithms~\cite{haustein2019object, toussaint2018differentiable}, geometric predicates are used to model the geometric relationship between the objects and the environment. Although these predicates are typically proposed in the context of known objects with geometric templates, they can benefit from shape completion and naturally generalize to a category of objects. Here we summarize several examples used in these manipulation planners. 

\begin{itemize}
    \item The static stability constraint enforces that the object placement surfaces are aligned with one of the environment placement regions. To use this predicate, it is required to extract the surfaces on the object that afford placing from the object dense geometry. Please refer to~\cite{haustein2019object} for more details.
    \item The visibility constraint requires the line segments from the sensor to the object are not blocked by other objects or the robot. In other words, the manipulate object should not block or be blocked by other objects.
    \item The containment constraint enforces the convex hull of an object is included in a container. This constraint needs the convex hull of the object, which can be computed using the dense geometry from shape completion. 
\end{itemize}

\subsection{Grasping}
\label{subsec:grasping}

The grasp planning module is responsible to compute a grasp pose that allows the robot to stably pick up and transfer the object. Various algorithms~\cite{morrison2018closing, gualtieri2016gpd, mahler2016dex} have been proposed for grasp planning. Some of them~\cite{varley2017shape, lundell2019robust, watkins2018multi} are built upon shape completion and can be easily integrated into our pipeline. These algorithms are object-agnostic and can robustly generalize to novel instances within a given category. 


\begin{figure*}[t]
\centering
\includegraphics[width=0.85\textwidth]{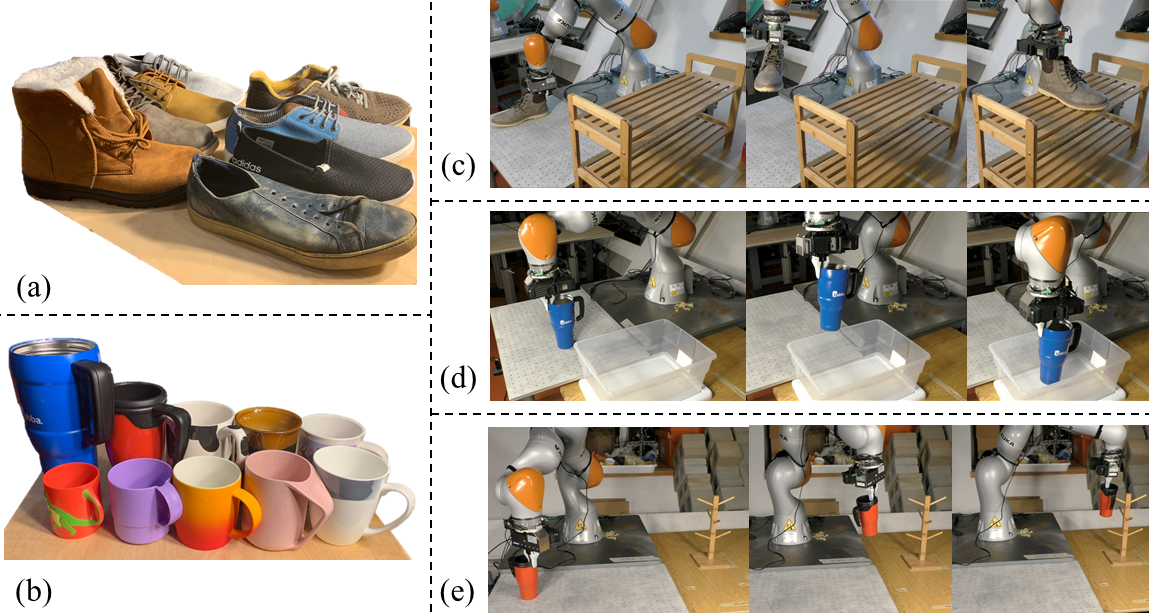}
\caption{\label{fig:result} An overview of our experiments. (a) and (b) are the shoes and mugs used to test the manipulation pipeline. Note that both the shoes and mugs contain substantial intra-category shape variation. We use three manipulation tasks to evaluate our method: (c) put shoes on a shelf; (d) put mugs in a container; (e) hang mugs on a rack by the mug handles. Readers are recommended to visit our \href{https://sites.google.com/view/generalizable-manipulation/home}{\textcolor{blue}{\underline{project page}}} to watch the video demo. }
\end{figure*}

\begin{figure}[t]
\centering
\includegraphics[width=0.45\textwidth]{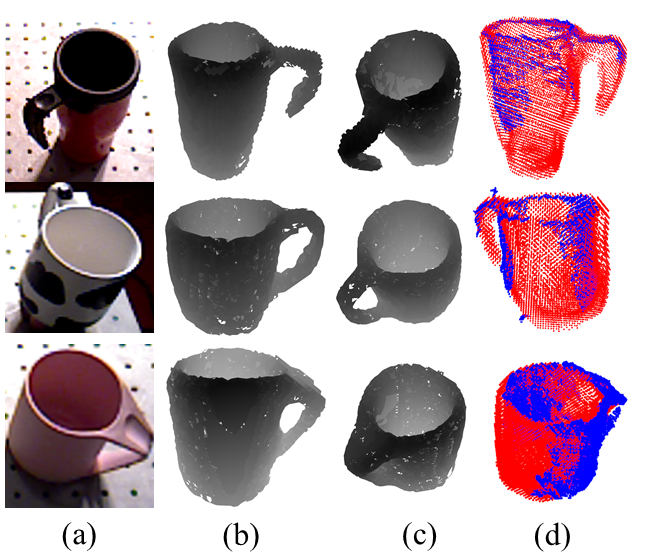}
\caption{\label{fig:perception_result} The qualitative results for shape completion. (a) is the input RGB image. (b) and (c) are the dense geometry from shape completion in two viewing directions. (d) compares the point cloud from depth image and shape completion: the depth image point cloud is in blue while shape completion is in red. Although the completed geometry contains small holes and defects, the accuracy is sufficient for many manipulation tasks. }
\end{figure}

\begin{figure}[t]
\centering
\includegraphics[width=0.45\textwidth]{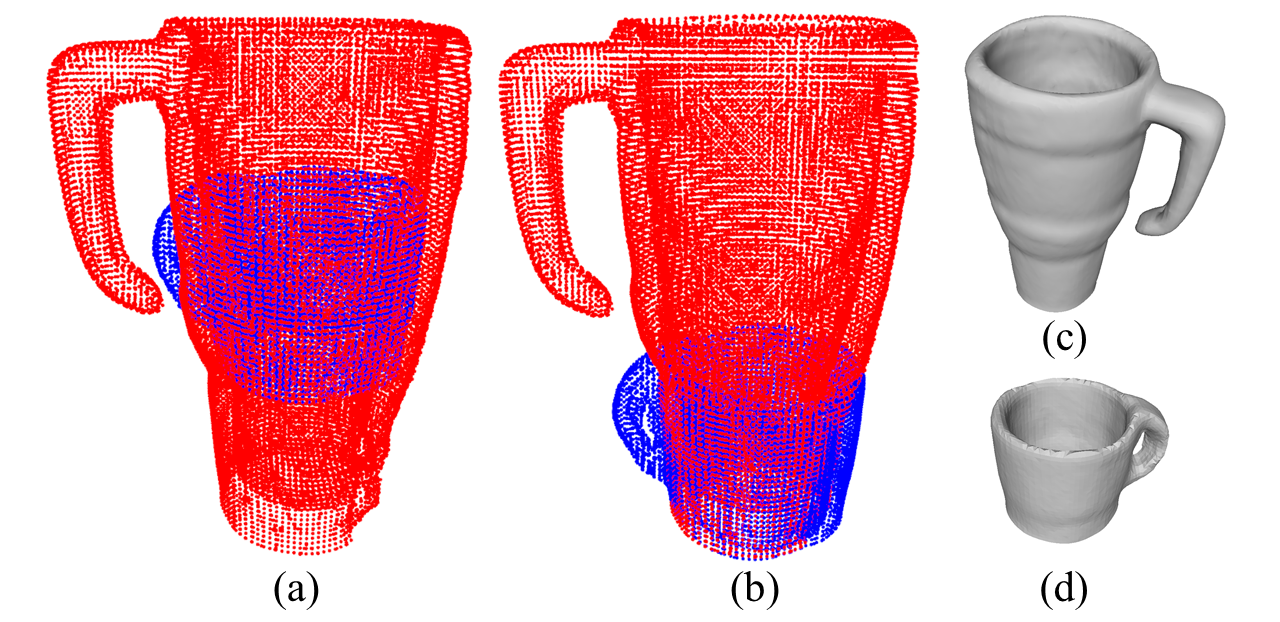}
\caption{\label{fig:pose_compare} Pose representation cannot capture large intra-category variations. (a) and (b) show two alignment results by \cite{gao2018filterreg} (variation on the random seed), where we attempt to register the mug template (d) into the observation (c). Using these pose estimation results for manipulation planning can lead to physically infeasibility, as the rigid transformation defined on the mug template (d) cannot capture the geometry of mug (c). Additionally, there may exist multiple suboptimal alignments which make the pose estimator ambiguous. }
\end{figure}

\section{Results}
\label{sec:results}

In this section, we demonstrate a variety of manipulation tasks using our pipeline. The particular novelty of these demonstrations is that our method can automatically plan robot trajectories that handle large intra-category variations without any instance-wise tuning or specification. 
The video demo and source code are available on our \href{https://sites.google.com/view/generalizable-manipulation/home}{\textcolor{blue}{\underline{project page}}}.

\subsection{Experiment Setup and Implementation Details}

We use 8 shoes and 10 mugs to test the manipulation pipeline, as illustrated in Fig.~\ref{fig:result}. Note that both shoes and mugs have substantial intra-category shape variation. The keypoints same as kPAM~\cite{manuelligao2019kpam} are used to define the target configuration of the shoe and mug. The statistics of the training objects is shown in Table \ref{table:training_data}.

We utilize a 7-DOF robot arm (Kuka IIWA LBR) mounted with a Schunk WSG 50 parallel jaw gripper. An RGBD sensor (Primesense Carmine 1.09) is also mounted on the end effector and used for all the perception tasks. Both the intrinsic and extrinsic parameters of the RGBD camera are calibrated.

\begin{table}[t]
  \caption{Training Data Statistics}
  \label{table:training_data}
  \includegraphics[width=\linewidth]{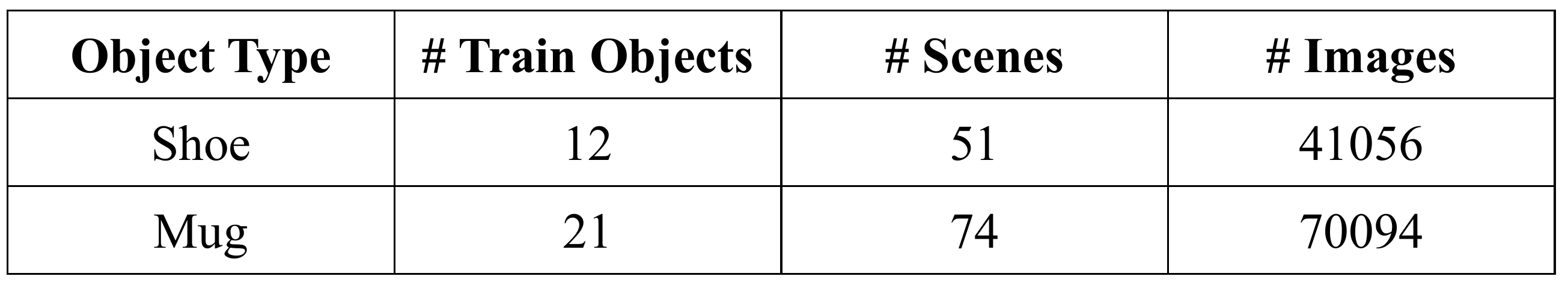}
\end{table}

\begin{table}[t]
  \caption{Robot Experiment Statistics}
  \label{table:success_rate}
  \includegraphics[width=\linewidth]{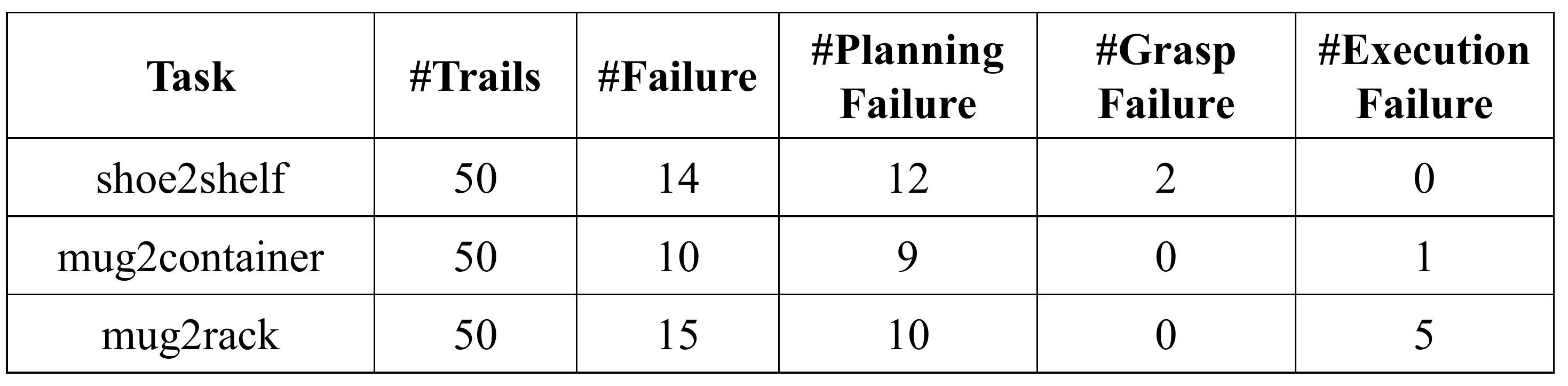}
\end{table}

The drake library~\cite{drake}, which implements optimization based motion planning, is used for manipulation planning. The costs and constraints include object target configuration and collision avoidance. For the purpose of this work, we use a fixed contact-mode sequence consists of picking, transferring and placing of the object, as the scheduling of contact-mode sequences is the main focus of task-level planning. We emphasize that the proposed object representation and manipulation pipeline are agnostic to the concrete planning algorithm that solves~(\ref{equ:problem}). Many motion planners, either optimization-based methods~\cite{schulman2013finding, toussaint2018differentiable} or sampling-based approaches~\cite{garrett2018sampling, srivastava2014combined}, can be plugged in and used.

\subsection{Perception and Comparison with Pose Estimation}
\label{subsec:sc_result}

In this subsection, we provide results of shape completion and compare it with the widely-used 6-DOF pose representation. Fig.~\ref{fig:perception_result} shows several completed dense geometry for representative mugs. The network takes input from images in Fig.~\ref{fig:perception_result} (a) and produces the dense geometries in Fig.~\ref{fig:perception_result} (b) and (c). Fig.~\ref{fig:perception_result} (d) compares the point cloud from depth images and shape completion. Although the completed geometry contains small holes and defects, the accuracy is sufficient for many manipulation tasks. Note that the network generalizes to instances with substantial variations on geometry and topology.

On the contrary, if object templates with pose estimation are used to obtain the dense geometry, as many existing works do, can result in multiple sub-optimal alignments that make the pose estimation ambiguous. An illustration is provided in Fig.~\ref{fig:pose_compare}. where the pose estimator in~\cite{gao2018filterreg} is used to align two mugs. Using these pose estimation results for manipulation planning can lead to physically infeasibility. Thus, pose estimation cannot generalize and is not suitable for the manipulation of a category of objects.

\subsection{Manipulation Task Specifications}

We use the following three manipulation tasks to test our pipeline:

\vspace{0.3em}
\noindent \textbf{1) Put shoes on a shelf: } The first manipulation task is to put shoes on a shoe shelf, as shown in Fig.~\ref{fig:result} (c). The final shoe configuration should be in alignment with the shelf. The manipulation pipeline has the pre-built template of the robot and the shoe rack, but need to deal with shoes with different appearance and geometry.

\vspace{0.3em}
\noindent \textbf{2) Place mugs into a container: } The second manipulation task is to place mugs into a box without colliding with it, as shown in Fig.~\ref{fig:result} (d). The mug should be upright and its handle should be aligned with the container.

\vspace{0.3em}
\noindent \textbf{3) Hang mugs on a rack: } The last manipulation task is to hang mugs onto a mug rack by the mug handle, as shown in Fig.~\ref{fig:result} (e). The geometry and position of the mug rack are available to the pipeline.
\vspace{0.3em}

Note that although task 1) and 3) have been performed in our previous work kPAM~\cite{manuelligao2019kpam}, it uses various manually-specified intermediate robot configurations to ensure physical feasibility. This manual specification is labor-intensive and can hardly scale to more complex environments and manipulation tasks. In contrast, with the integration of shape completion and manipulation planner, the physical feasibility is automatically ensured by the pipeline and the planned trajectories are much more efficient.

\subsection{Result and Failure Mode}

The video demo is on our \href{https://sites.google.com/view/generalizable-manipulation/home}{\textcolor{blue}{\underline{project page}}}. The statistics about three different tasks is summarized in Table.~\ref{table:success_rate}. Most failure cases result from the failure of the motion planner. Since the motion planner used in this work uses non-convex optimization internally, it can be trapped in bad local minima without a good initialization. This problem can be resolved by using sampling-based motion planners such as RRT or RRT*. These planners are globally optimal, although they need longer running time.

The grasp failure in Table.~\ref{table:success_rate} means (1) the robot fails to grasp the object; or (2) the relative motion between the gripper and the grasped object is too large. The relative motion may occur during the grasping, or when the object is not rigid. This problem could be alleviated by the addition of an external camera that would allow us to re-perceive the object after grasping.

The execution failure in Table.~\ref{table:success_rate} refers to the situations such that (1) the robot makes collision (despite the planning is successful); or (2) the object is not placed into the target configuration. This problem necessitates the execution monitoring described in~\cite{diankov2010automated}.


\section{Conclusion}
\label{sec:conclusion}

In this paper, we focus on manipulation planning of a category of objects, where the robot should move a set of objects to their target configuration while satisfying physical feasibility. This is challenging for existing manipulation planners as they assume known object templates and 6-DOF pose estimation, which doesn’t generalize of novel instances within the category. Thus, we propose a new hybrid object representation consists of semantic keypoints and dense geometry as the interface between the perception module and planning module. On the perspective of implementation, we contribute a novel integration of shape completion with keypoint detector and manipulation planner. In this way, both the perception and planning module generalizes to novel instance. Extensive hardware experiments demonstrate our manipulation pipeline is robust to large intra-category shape variation.

\section*{Acknowledgements}

The authors would like to thank Lucas Manuelli and Peter Florence for their help regarding robot setup and insightful suggestion. This work was supported by NSF Award IIS-1427050, Amazon Research Award and Draper Laboratory Incorporated Award No. SC001-0000001002. The views expressed in this paper are those of the authors themselves and are not endorsed by the funding agencies.

{\small
\bibliographystyle{abbrv}
\bibliography{paper.bib}
}

\end{document}